%% file: paper.tex
\documentclass[conference]{IEEEtran}
\IEEEoverridecommandlockouts
\usepackage{cite}
\usepackage{amsmath,amssymb,amsfonts}
\usepackage{algorithmic}
\usepackage{graphicx}
\usepackage{textcomp}
\usepackage{xcolor}
\def\BibTeX{{\rm B\kern-.05em{\sc i\kern-.025em b}\kern-.08em
		T\kern-.1667em\lower.7ex\hbox{E}\kern-.125em}}

\usepackage{times}
\usepackage{epsfig}

\usepackage{subcaption}
\usepackage[export]{adjustbox}
\graphicspath{{figures/}}
\usepackage[acronym]{glossaries}
\usepackage{bm}

\begin{document}
\input{text/header.tex}
\input{text/abstract.tex}
\input{text/intro.tex}
\input{text/data.tex}
\input{text/methods.tex}
\input{text/experiments.tex}
\input{text/end.tex}
{\small
\bibliographystyle{plain}
\bibliography{literature}
}
\end{document}

%% file: text/header.tex
\title{Semmeldetector: Application of Machine Learning in Commercial Bakeries\\
	\thanks{This study is partially supported by the European Social Funds (ESF) \mbox{No. R.6-V0332.2.43/1/5}.}
	%\thanks{This study is partially supported by someone (but this is a blind review).}
}
\author{
	\IEEEauthorblockN{
		Thomas H. Schmitt\IEEEauthorrefmark{1}, 
     	Maximilian Bundscherer and
		Tobias Bocklet
	}
	\IEEEauthorblockA{\textit{Department of Computer Science} - \textit{Technische Hochschule Nürnberg Georg Simon Ohm} Nuremberg, Germany}
	\IEEEauthorblockA{\IEEEauthorrefmark{1}Email: thomas.schmitt@th-nuernberg.de}
}
%
%\author{
%    	\IEEEauthorblockN{
%		  	Given Name Surname\IEEEauthorrefmark{1}, 
%			Given Name Surname and
%			Given Name Surname
%			}
%	    \IEEEauthorblockA{\textit{dept. name of organization (of Aff.)} - \textit{name of organization (of Aff.)} City, Country}
%		\IEEEauthorblockA{\IEEEauthorrefmark{1}Email address or ORCID}
%}
%
\maketitle

%% file: text/abstract.tex
\begin{abstract}
The Semmeldetector, is a machine learning application that utilizes object detection models to detect, classify and count baked goods in images.
Our application allows commercial bakers to track unsold baked goods, which allows them to optimize production and increase resource efficiency.
We compiled a dataset comprising \(1151\) images that distinguishes between \(18\) different types of baked goods to train our detection models.
To facilitate model training, we used a Copy-Paste augmentation pipeline to expand our dataset.
We trained the state-of-the-art object detection model YOLOv8 on our detection task.
We tested the impact of different training data, model scale, and online image augmentation pipelines on model performance.
Our overall best performing model, achieved an \(AP_{0.5}\) of \(89.1\%\) on our test set.
Based on our results, we conclude that machine learning can be a valuable tool even for unforeseen industries like bakeries, even with very limited datasets.
\end{abstract}
\glsresetall

%% file: text/intro.tex
\section{Introduction}
\label{sec:intro}
The Semmeldetector, named after the locally used German word for bread bun, is a machine learning application that utilizes state-of-the-art object detection model YOLOv8 \cite{YOLOv8} to detect, classify, and count baked goods in images.
However, due to the vast diversity of baked goods in Germany, with each bakery offering its unique assortment, to the best of our knowledge, there are no datasets available that sufficiently differentiate between baked goods.
To train our models, we created a dataset comprising \(1151\) images distinguishing between \(18\) types of baked goods.
We utilized SAM \cite{SAM} to annotate our training data to streamline and speed up the annotation process.
To facilitate model training, we employed a Copy-Paste augmentation \cite{CopyPaste} pipeline to expand our training data.
%
%\newline
%\newline
%
Our object detection models allows commercial bakers to automatically track unsold baked goods, optimizing production, increasing resource efficiency, and meeting industry partner requirements.
Which eliminates the often costly manual tracking that would otherwise be required.
%
%Baked goods are typically convex objects that fit well within their respective bounding boxes, making object detection models a suitable choice for our application.
%
%Compared to e.g., image segmentation, object detection offers the additional advantage of being more robust against localization errors.
%
%Object detection models can maintain accurate counts even with localization errors, while image segmentation models may exhibit chaotic (small errors can significantly impact count predictions) behaviour, with respect to localization errors.
%
%\newline
%\newline
%
The main contributions of this study are:
(1) The application of computer vision models in commercial bakeries to unsold product.
(2) The demonstration of the effectiveness of the Copy-Paste augmentation \cite{CopyPaste} to enrich small datasets.
(3) The deployment of our models as an iOS application, offering commercial bakeries a user-friendly platform to easily utilize our models.
\subsection{Related Work} % Tobias: Aufteilung von Related Work nicht nötig
\label{sec:sec:related_work}
%
% Dominik: Erwähnen das sehr Limited um Wert eigene Arbeit hervorzuheben.
%
%\subsubsection{Object detection on or in baked goods}
%\label{sec:sec:sec:detection_on_in_bread}
%
Application studies \cite{RelateWork_detection_in_bread_1} and \cite{RelateWork_detection_in_bread_2} used U-net \cite{U-net} and YOLOv5 \cite{YOLOv5} models to detect defects on or in baked goods, respectively.
Both studies aimed at improving food safety using machine learning.
The first utilized a combination of near infrared (NIR) spectroscopy images and computer vision to detect foreign contaminants in toast bread.
The second used image data to detect mold on the surface of various food items, including baked goods.
Both studies achieved detection accuracies of over \(95\%\).
These results demonstrate the effectiveness of machine learning in improving food safety.
\newline
\newline
%
%\subsubsection{Detection of baked goods}
%\label{sec:sec:sec:detection_of_bread}
%
Application study \cite{RelateWork_detection_of_bread_3} employed a SVM to perform image segmentation on images of a specific kind of flatbread.
Their primary objective was to ensure quality control by accurately and quickly predicting the size and shape of bread sheets in various scenarios.
To achieve this, they operated color-based in a relatively controlled image environment.
They achieved a maximum error rate of \(2.2\%\).
Application studies \cite{RelateWork_detection_of_bread_1} and \cite{RelateWork_detection_of_bread_2} used image processing models to detect baked goods in images.
The goal of \cite{RelateWork_detection_of_bread_1} was to automate visual quality inspection during the bread production process, while \cite{RelateWork_detection_of_bread_2} aimed at detecting baked goods in images.

%% file: text/data.tex
\section{Data}
\label{sec:data}
\begin{figure}
	\begin{center}
		\includegraphics[trim={0 0.25cm 0 0.25cm},clip,width=0.98\linewidth]{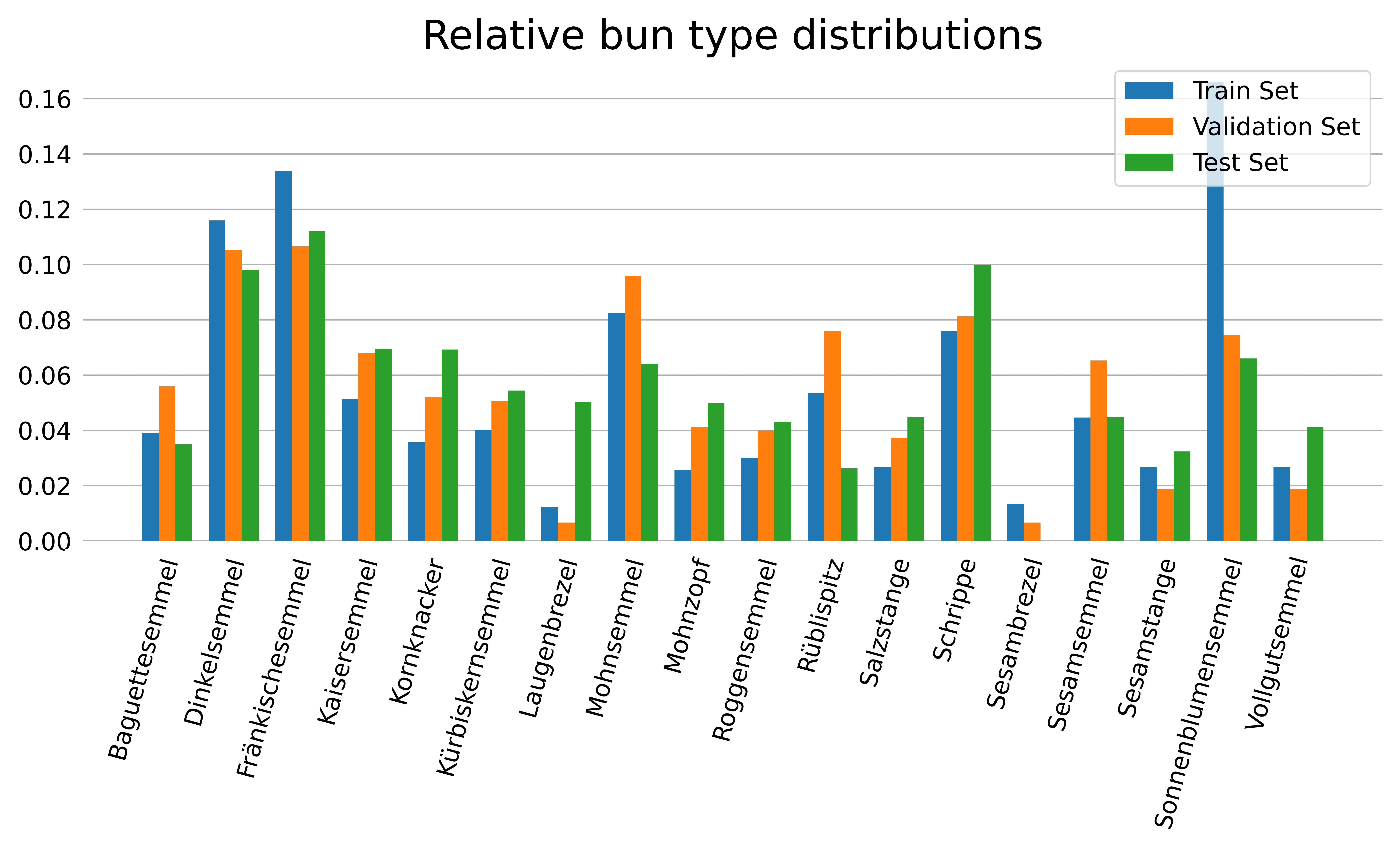}
	\end{center}
	\caption{Relative baked good distributions in our baseline training, validation and test set.} % Grafik erklären
	\label{fig:histo}
\end{figure}

Our baseline dataset comprises \(1151\) images of baked goods divided into \(897\) training, \(45\) validation, and \(209\) test images.
We distinguished between \(18\) different types of baked goods.
Some of which, such as Sonnenblumensemmel (sunflower bread bun) and Vollgutsemmel (wholemeal bread bun), can be difficult to distinguish even for human annotators (see top right image in Figure \ref{fig:aug_examples}).
We captured images in a relatively controlled environment that closely resembles our use case.
%
% A metal drying tray served as the background for our images to further simulate the use case of our models.
%
The training and validation set images were taken with HD webcams, while the test set images were taken with an iPad.
The relative distribution of baked goods in our base datasets is shown in Figure \ref{fig:histo}.
%
% Max: vielleicht darauf eingehen woher die relative distribution her kommt.
%
\subsection{Training Set}
\label{sec:sec:train_set}
Our training set comprises \(897\) images of baked goods captured from two fixed camera angles.
The camera angles used were directly from above and at a slight angle to simulate our models optimal operating conditions.
%
% Tobias: Bild von Setup
%
We limited our training set to images featuring a single baked good, positioned roughly in the centre. 
While our imposed constraint greatly diminish the informative value of our training set images, which we counteract with data augmentation (Section \ref{sec:sec:image_composition}), it grants us the following advantages:
%
% Max: voriger Satz unklar
%
(1) It allowed us to automatically annotate our training set images using a SAM annotation pipeline (Section \ref{sec:sec:image_annotation}).
(2) It allowed us to focus the training set on challenging out-of-plane rotations that are hard to simulate.
(3) It allows us to rapidly scale our models to new baked goods thanks to the automated annotation process.
\subsection{Validation Set}
\label{sec:sec:validation_set}
Our validation set comprises \(45\) images of baked goods,
with an average of \(17\) baked goods per image.
The images in our validation set were captured directly from above, and although our training and validation sets are disjoint, the same baked good samples were used to form both of them.
%
% Dominik: Wird das beim Testing nicht gemacht
%
To evaluate our model's generalization capabilities to unknown scenarios 
we designed our validation set with a diversity of arrangements and orientations,
while keeping the overlap between baked goods to a minimum.
To validate the robustness of our models against varying lighting conditions, we varied the lighting conditions in our validation set images.
To enrich our validation set and validate the robustness of our models against image scale and rotation, we derived \(1000\) supplementary validation set images by rotating and scaling our base images.
We used the resulting validation set \(val\) comprising \(1045\) images, to evaluate our models during training.
\subsection{Test Set}
\label{sec:sec:test_test_set}
Our test set comprises \(209\) images of baked goods,
with an average of \(15\) baked goods per image.
We form our test set with new baked good samples collected on a seperate day,
which allows us to test whether our models are overly adapted to the particular baked goods present in our training and validation set.
Sesambrezeln (sesame pretzels) were excluded from our test set due to limited availability.
Our test set comprises various unseen use cases, enabling us to conduct a targeted evaluation of model behaviour.
To simplify analysis, the test set is divided into two primary subsets:
(1) \(test_{r}\): \(74\) images taken from various camera angles to test model performance in non-ideal deployment conditions.
(2) \(test_{i}\): \(133\) images taken from a fixed, optimal angle. 
Subset \(test_{i}\) comprises \(test_{u}\): \(24\) images closely simulating our use case, \(test_{l}\): \(84\) images with exaggerated lighting conditions, and \(test_{c}\): \(27\) images of broken or crumbled baked goods.
We report model performances on the full test set, subsequently referred to as \(test\), and its two primary subsets \(test_{r}\) and \(test_{i}\).
Significant performance differences between secondary test subsets are reported.
\subsection{Image Annotation}
\label{sec:sec:image_annotation}
Validation and test set images were manually annotated.
A baked good was annotated with a bounding box if at least \(20\%\) of it was visible.
Despite the frequent baked good occlusion in our use case, it is reasonable to assume that they are at least \(20\%\) visible, due to the subsequent processing process.
Our application is designed to count baked goods, which is why we annotate baked goods that were split due to overlap by others with a single extrapolated bounding box if at least \(10\%\) of it is visible on both sides.
%
% Otherwise, only the larger portion of the baked good is annotated with a bounding box.
%
%\newline
%\newline
%
% Max: sprung warum training set images automatisch annotiert werden können unklar.
%
Training set images were automatically annotated using the SAM annotation pipeline.
Given our controlled image enviroment, annotations are derived from the largest non-background segmentation mask in each image.
A segmentation mask is considered to belong to the background if its corresponding bounding box has an \(IoU\) of at least \(90\%\) with the image.
We utilized morphological operations to remove small isolated regions and holes in the segmentation masks.

%% file: text/methods.tex
\section{Methods}
\label{sec:methods}
\subsection{Copy-Paste Augmentation}
\label{sec:sec:image_composition}
\begin{figure}
	\centering % <-- added
	\begin{subfigure}{0.48\linewidth}
		\includegraphics[width=\linewidth, trim={2.5cm 0 2.5cm 0},clip]{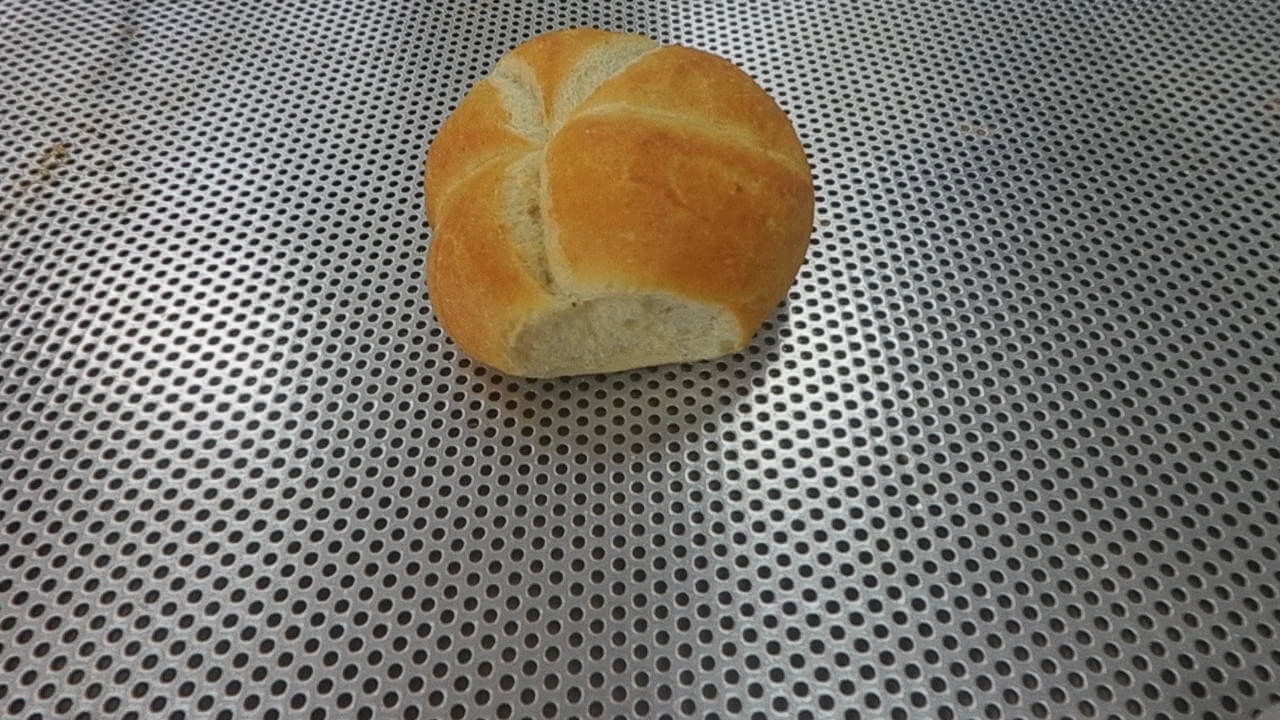}
		\label{fig:examples_1}
	\end{subfigure}\hfil % <-- added
	\begin{subfigure}{0.48\linewidth}
		\includegraphics[width=\linewidth, trim={5cm 0 5cm 0},clip]{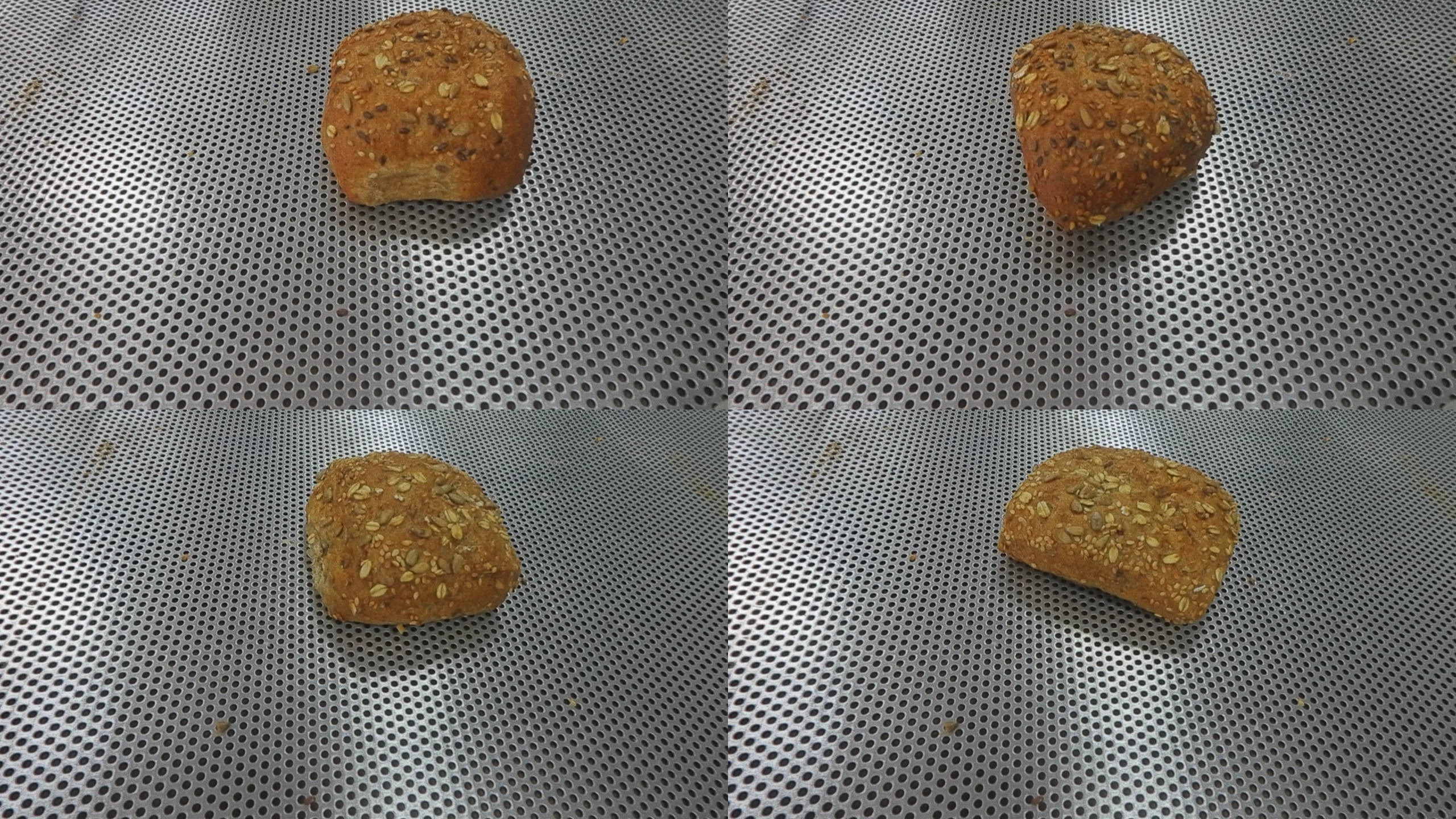}
		\label{fig:examples_2}
	\end{subfigure}
	\medskip
	\begin{subfigure}{0.48\linewidth}
		\vspace{-0.3cm}
		\includegraphics[width=\linewidth, trim={5cm 0 5cm 0},clip]{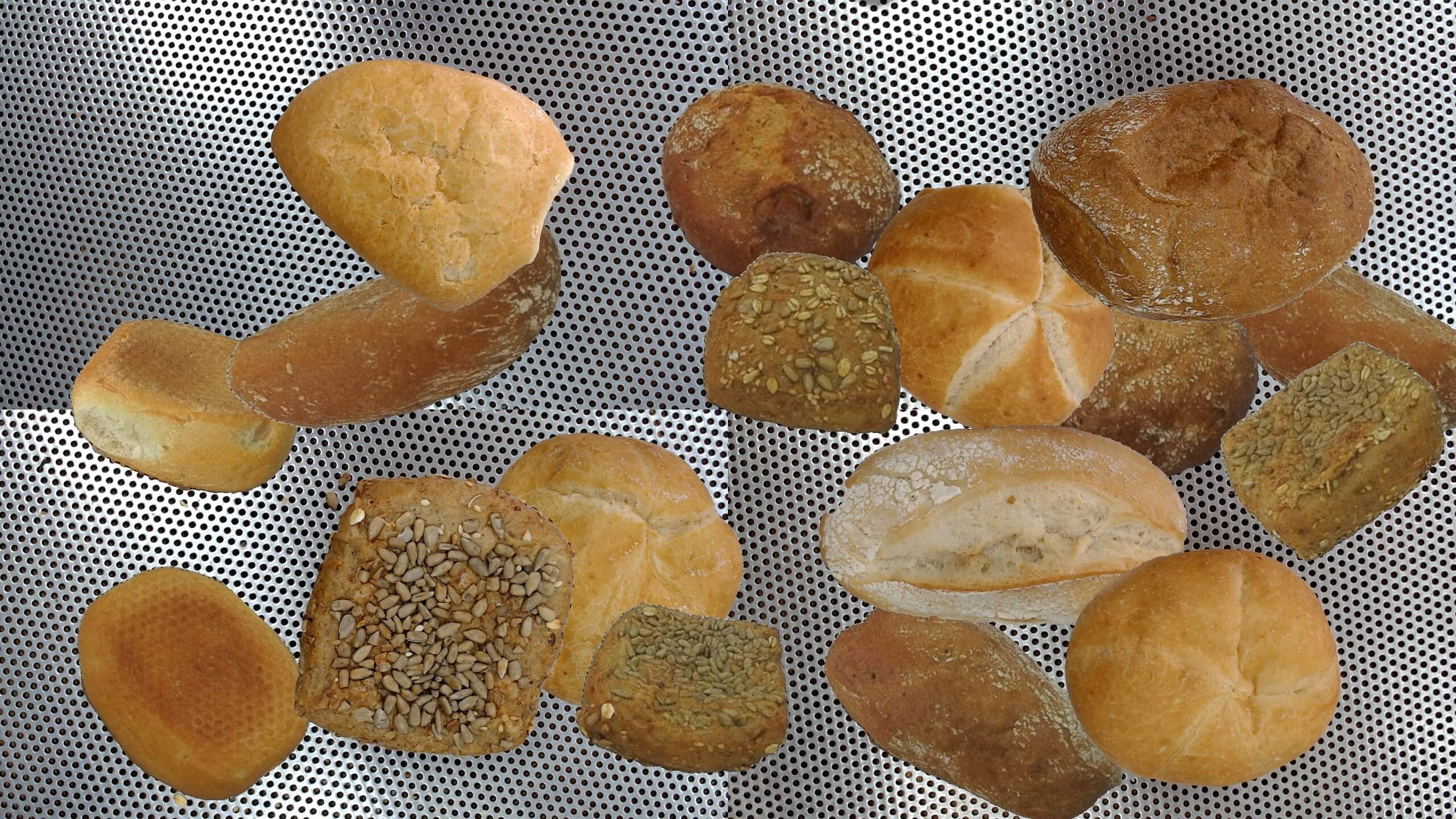}
		\label{fig:examples_3}
	\end{subfigure}\hfil % <-- added
	\begin{subfigure}{0.48\linewidth}
		\vspace{-0.3cm}
		\includegraphics[width=\linewidth, trim={2cm 0 2cm 0},clip]{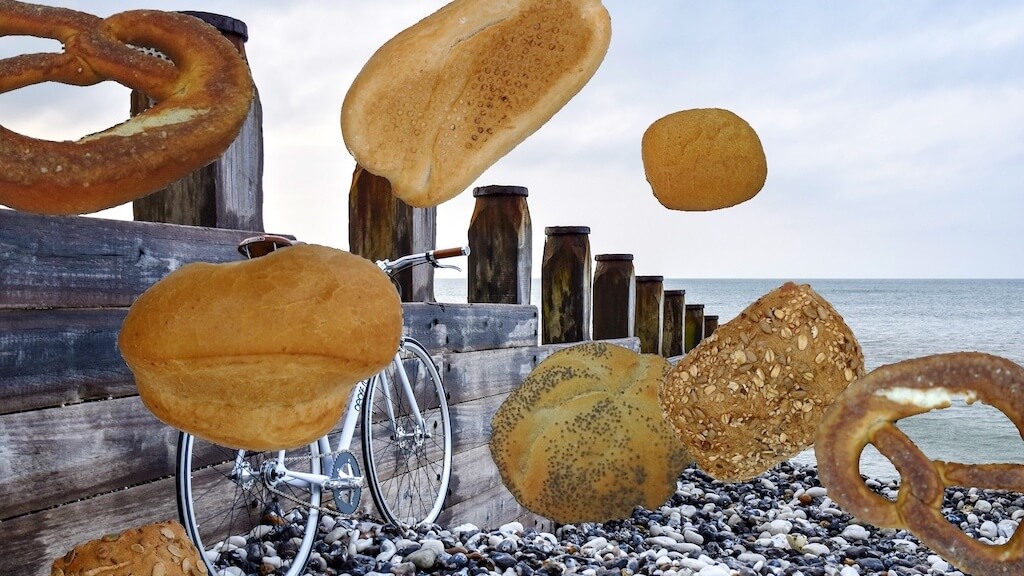}
		\label{fig:examples_4}
	\end{subfigure}
	\medskip
	\begin{subfigure}{0.48\linewidth}
		\vspace{-0.52cm}
		\includegraphics[width=\linewidth, trim={5cm 0 5cm 0},clip]{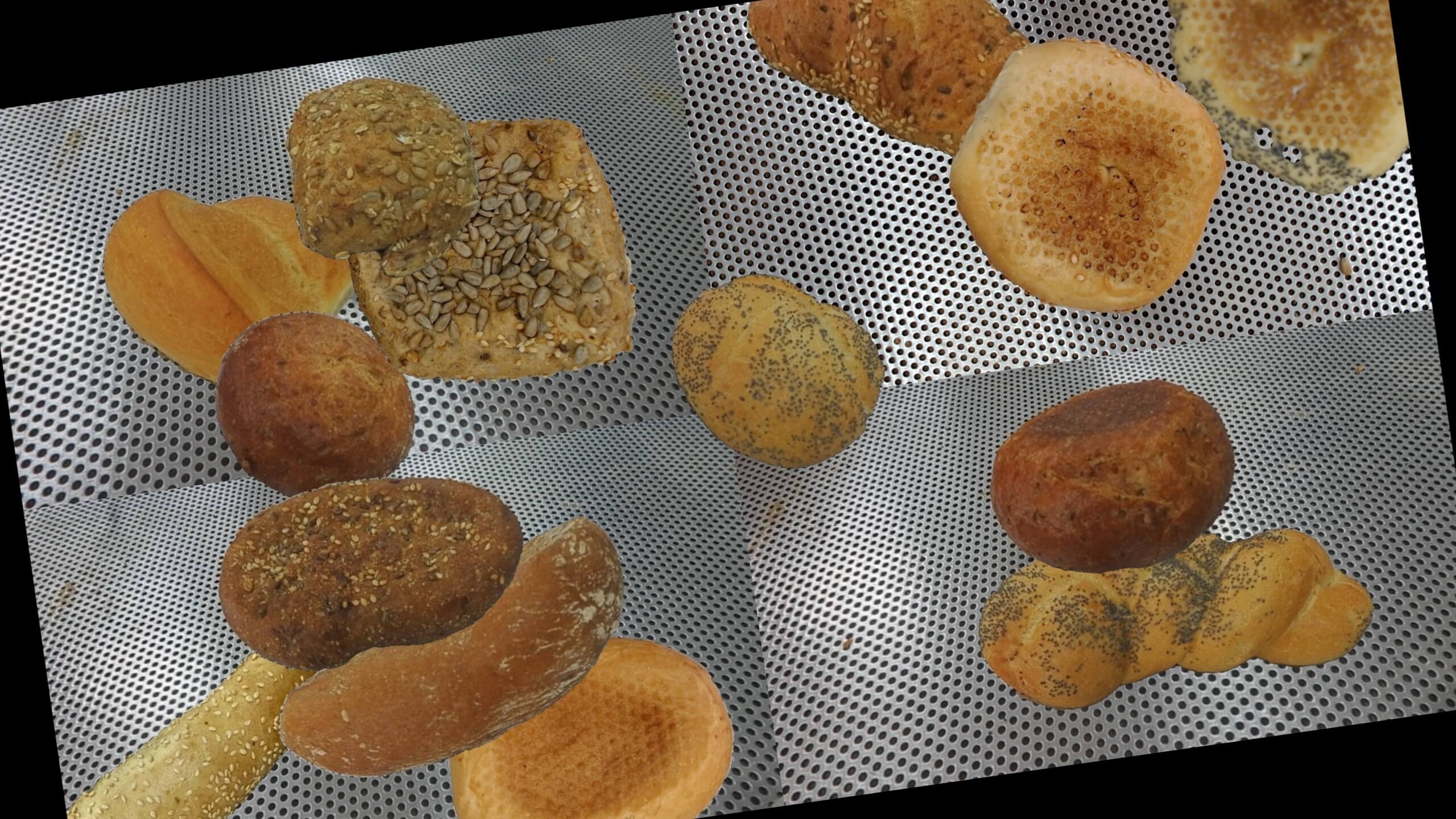}
		\label{fig:examples_5}
	\end{subfigure}\hfil % <-- added
	\begin{subfigure}{0.48\linewidth}
		\vspace{-0.52cm}
		\includegraphics[width=\linewidth, trim={5cm 0 5cm 0},clip]{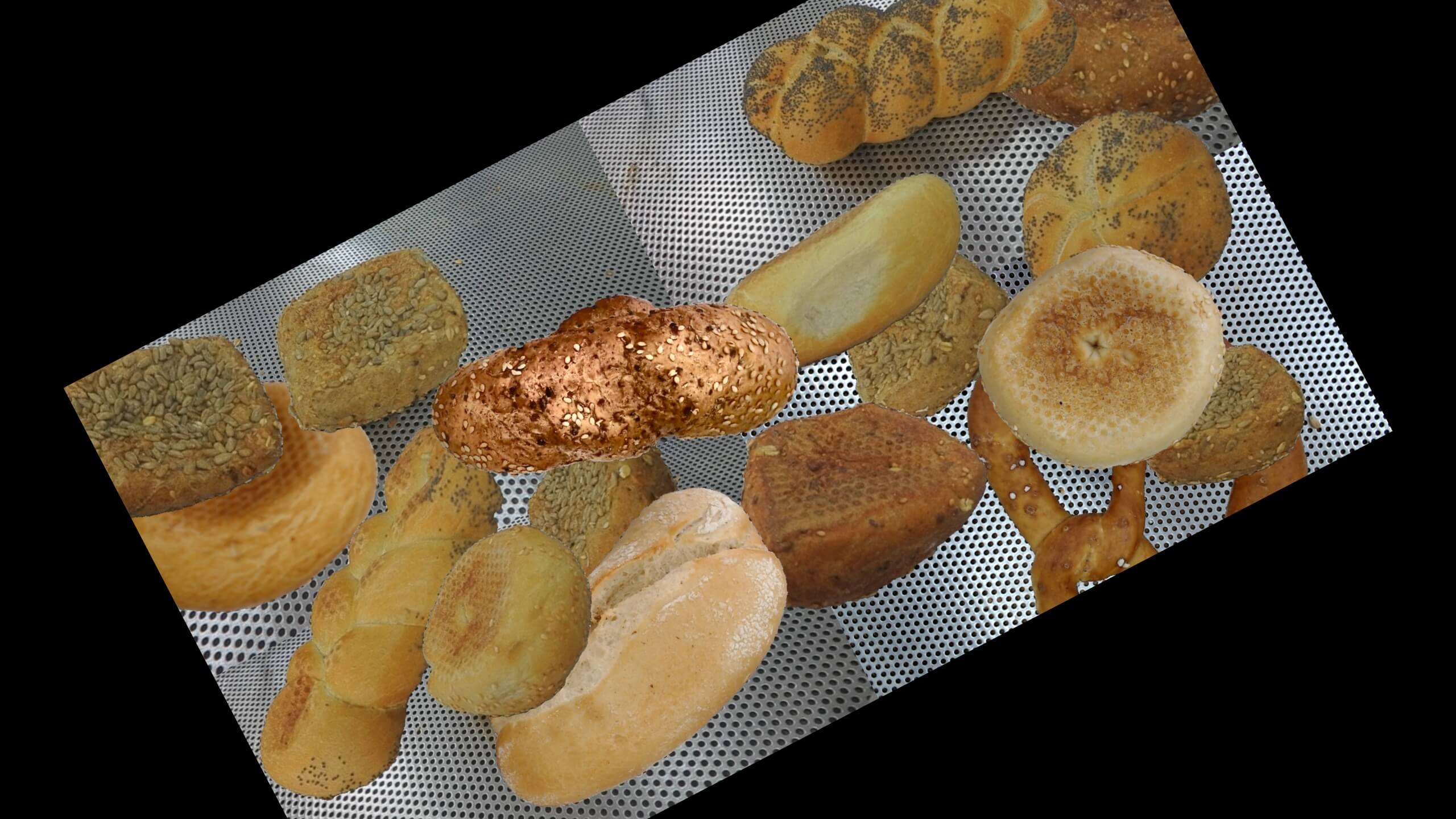}
		\label{fig:examples_6}
	\end{subfigure}
	\caption{Training set images: (top left) image of a baked good, (top right) synthetic image featuring Sonnenblumensemmeln (sunflower bread buns) and Vollgutsemmeln (wholemeal bread buns), (middle left) synthetic image of baked goods, (middle right) synthetic image of baked goods on a beach, (bottom row) scaled and rotated synthetic baked good images.}
	\label{fig:examples}
\end{figure}

Our goal of automatically annotating our training data and rapidly scaling our models limited our initial training set to images of individual baked goods, likely rendering it inadequate for training large object detection models.
To counteract this limitation, we employed a Copy-Paste augmentation pipeline,
which combines information from multiple images by selectively copying the precise pixels corresponding to an object, rather than all pixels within the object's bounding box.
This object-aware approach allows the Copy-Paste augmentation to simulate occlusion interactions between objects, unlike mixing image augmentations like CutMix \cite{CutMix}.
\newline
%\newline
%
Using the Copy-Paste augmentation we iteratively created crowded baked good images, with an average of \(16\) baked goods per image.
%
%Baked goods were placed on free spots in the image, determined by the segmentation masks of the baked goods already present in the image.
%
%To limit complete occlusion of baked goods, we used the morphology dilation operation on the segmentation masks of the preceding image to maintain additional spacing between baked goods.
%
Bounding boxes of heavily occluded baked goods (less than \(10\%\) visible) were removed from the image annotations.
A range of augmentations were applied to the baked goods before pasting them into a image, namely: rotation, scaling, low-probability blur and CLAHE \cite{CLAHE}.
Similarly to \cite{CopyPaste}, we found that blending of pasted objects does not significantly impact model performance, we therefore opted not to blend pasted objects.
Background images were generated using a simplified version of the mosaic data augmentation method introduced in \cite{YOLOv4}.
To increase robustness against \(FP\), we used a subset of the DIV2K dataset \cite{DIV2K} as background images for synthetic images.
We evenly split the DIV2K dataset into two subsets: one for image synthesis, and one to evaluate our models.
We used the DIV2K dataset \cite{DIV2K} due to its higher image resolution compared to datasets like Microsoft COCO \cite{Coco}.
\newline
%\newline
%
We created a total of \(4000\) synthetic images of baked goods, with \(2000\) featuring a synthetic image background and \(2000\) featuring a random background sourced from DIV2K.
To enrich our training set and avoid overadapting our models to the grid structure introduced by the background generation, we created an additional \(2000\) images by scaling and rotating our synthetic images.
Our final training set comprises \(6897\) images, divided into three subsets:
(1) \(train_{b}\): the \(897\) baseline images.
(2) \(train_{s}\): \(4000\) synthetic images with a synthetic background.
(3) \(train_{n}\): \(2000\) synthetic images with a negative background sourced from DIV2K.
Some example images are shown in Figure \ref{fig:examples}.
\subsection{Online Image Augmentation}
\label{sec:sec:online_augmentation}
\begin{figure}
	\centering % <-- added
	\begin{subfigure}{0.48\linewidth}
		\includegraphics[width=\linewidth, trim={2.5cm 0 2.5cm 0},clip]{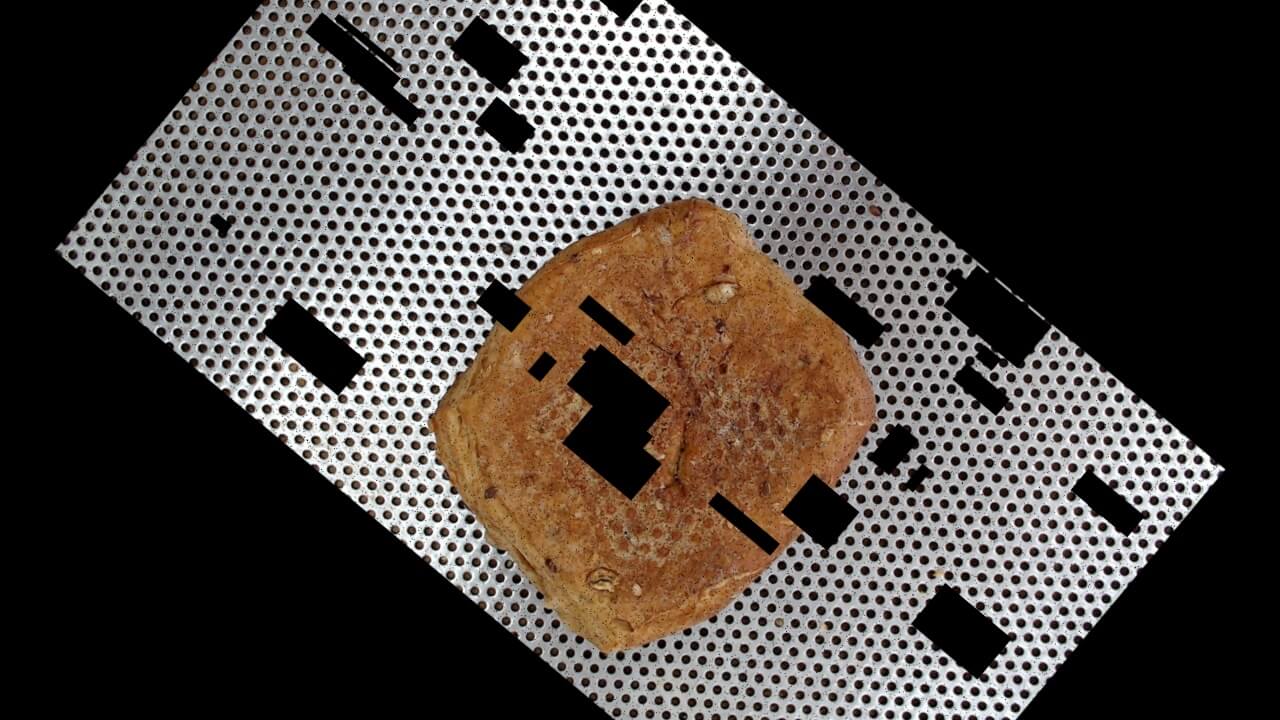}
		\label{fig:aug_examples_1}
	\end{subfigure}\hfil % <-- added
	\begin{subfigure}{0.48\linewidth}
		\includegraphics[width=\linewidth, trim={5cm 0 5cm 0},clip]{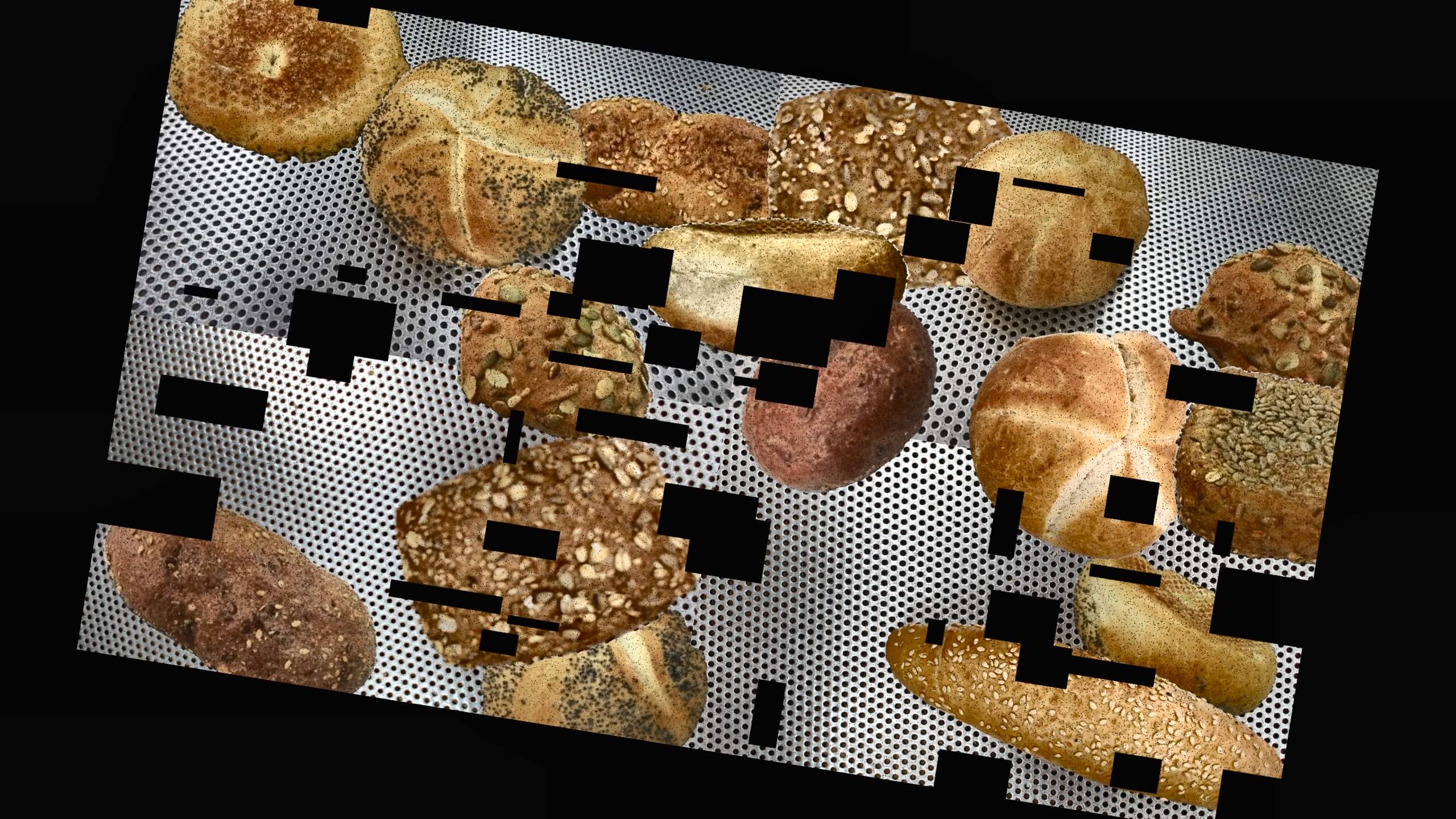}
		\label{fig:aug_examples_2}
	\end{subfigure}
	\medskip
	\begin{subfigure}{0.48\linewidth}
		\vspace{-0.3cm}
		\includegraphics[width=\linewidth, trim={5cm 0 5cm 0},clip]{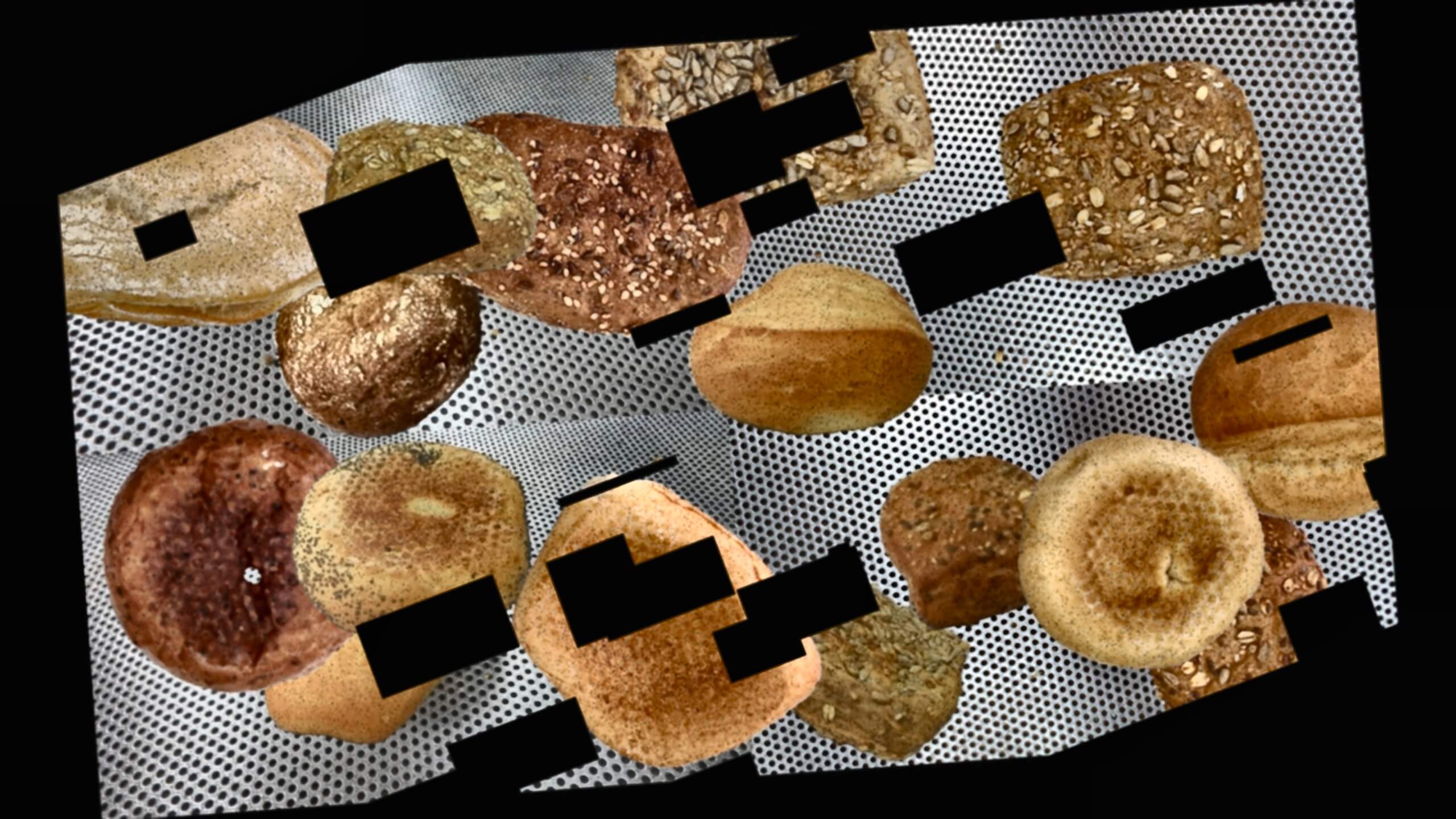}
		\label{fig:aug_examples_3}
	\end{subfigure}\hfil % <-- added
	\begin{subfigure}{0.48\linewidth}
		\vspace{-0.3cm}
		\includegraphics[width=\linewidth, trim={4cm 0 4cm 0},clip]{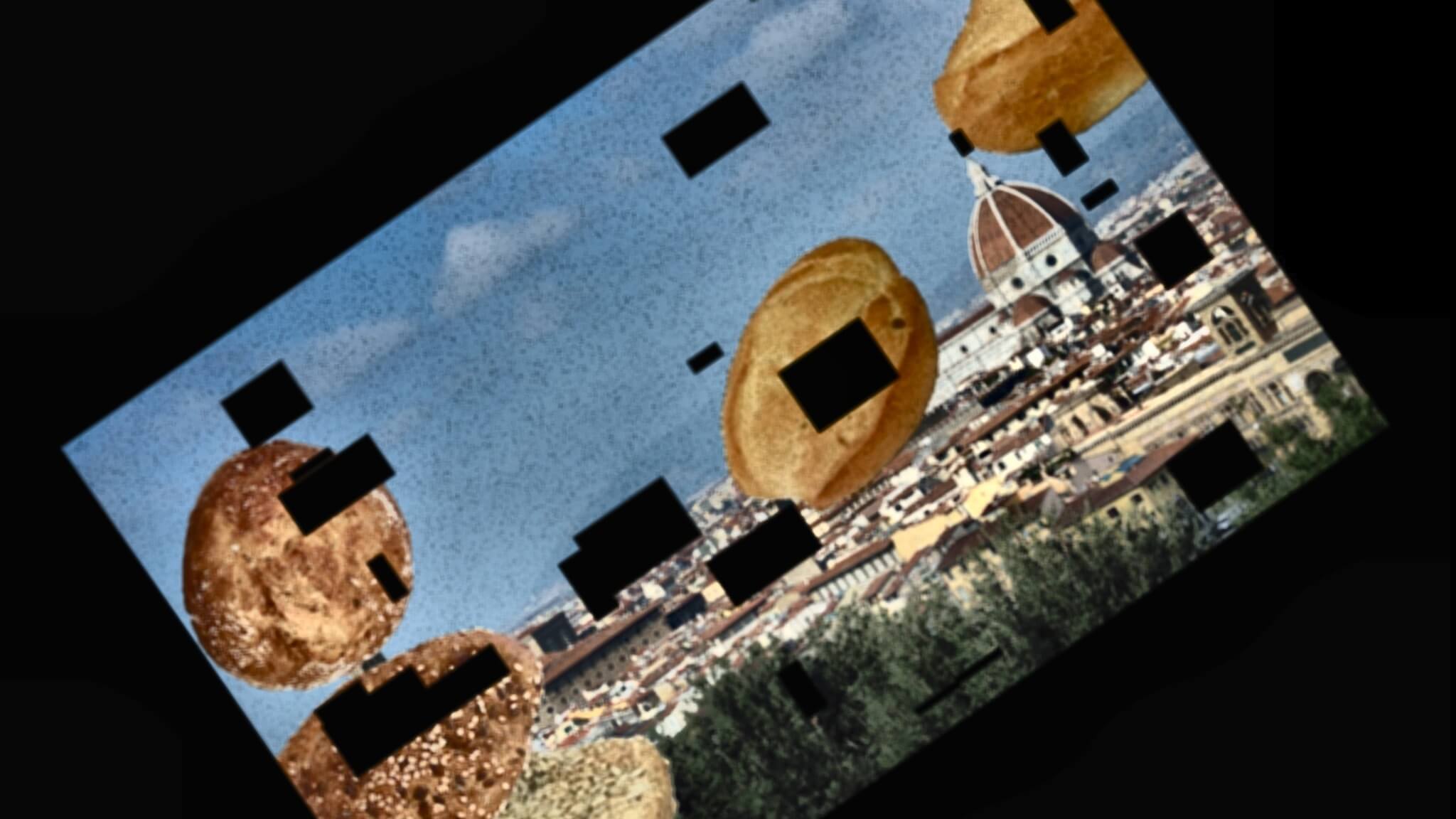}
		\label{fig:aug_examples_4}
	\end{subfigure}
	\caption{Examples of images after applying our online augmentation \(DO_{0.04}\).}
	\label{fig:aug_examples}
\end{figure}

In addition to our offline image augmentations (Section \ref{sec:sec:image_composition}), and our model's respective default online image augmentation pipelines, we employed two supplementary online image augmentation pipelines based on the Albumentations library \cite{Albu} to further facilitate model training.
Our online augmentation pipelines focused on simulating common image distortions occurring in our use case: 
out-of-focus baked goods, 
challenging lighting conditions, 
partially occluded or crumbled baked goods, 
and varying image scales and rotations.
Although most of these scenarios were already addressed in our image synthesis, we used online image augmentation to further improve model robustness.
\newline
%\newline
%
%Unlike \cite{Online_Augment}, which applies a random sub-policy consisting of two image augmentation operations, we applied our operations in a sequential manner.
%
Our baseline augmentation pipeline, denoted as \(BL_{0.01}\), applies pixel-level transformations.
The augmentations include, in order: Blur, MedianBlur, ToGray, and CLAHE \cite{CLAHE}, each applied with probability of \(0.01\).
To increase model robustness to scale, rotation, and baked good occlusion, we defined dropout augmentation pipeline \(DO_{0.04}\) that applies spatial and pixel-level transformations, followed by our base augmentation pipeline applied with a lower probability.
The augmentations include, in order: CoarseDropout, PixelDropout, Scale, Rotate.
%
%The additional transformations were applied with a higher probability due to their less destructive nature.
%
CoarseDropout was limited to areas comprising at most \(10\%\) of the image, therefore bounding boxes were assumed to remain unchanged.
Examples of images after applying \(DO_{0.04}\) are shown in Figure \ref{fig:aug_examples}.
\subsection{Tested Object Detection Models}
\label{sec:sec:models}
\subsubsection{Yolo}
\label{sec:sec:sec:yolo}
The YOLO (You Only Look Once) \cite{YOLOv1} object detection series, predicts bounding boxes and class labels in a single prediction step.
%
%YOLO models combine prediction of bounding boxes and class labels by framing object detection as a regression problem.
%
%This is achieved by dividing the input image into a grid, and predicting bounding box coordinates and class likelihoods for each grid cell separately.
%
This allows the model to detect objects with a single network, as opposed to two-stage object detection models like the R-CNN \cite{R-CNN} series that requires elaborate detection pipelines.
%
%The unified detection model is trained to optimize a weighted sum of a specialized classification loss and bounding box regression loss.
%
%During inference, YOLO models use non-maximal suppression to prevent duplicate detections of objects, which is especially important for maintaining a correct count of objects in an image.
%
YOLOv5 \cite{YOLOv5} and its successor YOLOv8 \cite{YOLOv8} have been designed to facilitate seamless transfer of the YOLO model to a wide range of object detection tasks. 
\subsubsection{defDETR}
\label{sec:sec:sec:deformable_detr}
The DETR (Detection Transformer) model \cite{DETR} utilizes the combination of a CNN backbone and a transformer encoder-decoder \cite{Transformer} to perform object detection,
by treating object detection as a direct prediction problem.
%
%DETR's streamlined end-to-end object detection pipeline removes the need for many hand-designed components like the non-maximum suppression used in YOLO models.
%
%Removing non-maximum suppression can improve Recall by preventing the rejection of neighbouring bounding box predictions in crowded images.
%
%DETR models prevent duplicate object detection by utilizing a bipartite matching loss to enforce unique object detection matchings.
%
%\newline
%\newline
%
defDETR (deformable DETR) \cite{defDETR} proposes improvements to the DETR model by introducing the deformable attention mechanism to improve both convergence speed and model performance.
The deformable attention mechanism mitigates the slow convergence and high complexity of the attention mechanism by combining attention with the sparse spatial sampling of deformable convolution \cite{defCNN}.
%
%Restricting the attention mechanism to a small fixed number of key sampling points reduces its complexity from quadratic to linear with respect to the feature map size.
%
Reducing the complexity allows the incorporation of higher resolution feature maps, improving model performance, especially when detecting small objects.
%

%% file: text/experiments.tex
%\newpage
\section{Experiments and Results}
\label{sec:experiments}
\subsection{Average Precision (AP)}
\label{sec:sec:average_precision}
Mean Average Precision \(mAP\), also commonly referred to as Average Precision \(AP\), is the primary metric used to evaluate the performance of object detection models.
%
%Dominik: machen die anderen Semmelerkenner das auch so?
%
It is calculated in relation to a \(IoU\) (Intersection over Union) threshold,
which is used to evaluate if a bounding box prediction is sufficiently overlapping with the ground truth.
After identifying bounding box predictions that detect a ground truth object, the \(AP\) for each class is determined by calculated the area under the corresponding \(PR\) (Precision-Recall) curve.
\(AP\) is the mean of these class-specific \(AP\)s.
Since our application places less emphasis on localization accuracy, we primarily use \(AP_{50}\) (\(AP\) at \(IoU\) threshold of \(50\%\)) to evaluate model performance.
The linear interpolation of the \(PR\) curve makes the \(AP\) sensitive to the number of threshold used for calculation.
Upper bound for relevant threshold values when calculating the \(PR\) curve is the total number of unique class likelihood predictions.
Cantor's theorem entails that the number of unique class likelihood predictions monotonically increases with the number of predictions.
\(AP\) is thus likely dependent on the number of predictions and not countably additive, e.g. \(AP_{50}^{test}\) is not necessarily equal to the weighted mean of \(AP_{50}^{test_r}\) and \(AP_{50}^{test_i}\).
\subsection{Experimental Setups} % Tobias: Vielleicht als Einleitung für Experiments and Results
\label{sec:sec:experimental_setup}
To improve model performance on our specialized task with its limited training data, we used models pre-trained on the COCO dataset.
We adapted and reinitialized model classification heads to suit our application.
Models were trained with their respective default training hyperparameters and augmentation pipelines,
with the addition of our online image augmentation pipeline.
%
%Models were trained using the stochastic gradient descent (SGD) optimizer with momentum and a learning rate cosine decay.
%
%Dominik: Warum nicht Adam/AdamW?
%
To guarantee convergence, models were trained for \(150\) epochs. 
%
%To guarantee convergence, models were trained for \(150\) epochs with a patience of \(50\) based on validation set performance.
%
We reported the performances of the final models, as opposed to the best performing models on the validation set.
%
% Significant performance differences between the final and best performing models are reported.
%
\subsection{Ablation Study}
\label{sec:ablation_study}
The baseline model for our ablation study was formed by:
training a large-scaled YOLOv8 model (YOLOv8-L), 
on our entire training set as grayscale images with a maximum image length of \(640px\) \(train_{b,s,n}^{gray}\), 
utilizing our dropout image augmentation pipeline \(DO_{0.04}\).
\subsubsection{Variations of Training Data}
\label{sec:sec:ablation_data}
To validate the effectiveness of our image synthesis and the focus on grayscale images, we trained our baseline model on:
our baseline training set \(train_b\),
our training set without images with random backgrounds \(train_{b,s}\),
and our entire training set \(train_{b,s,n}\),
either unchanged \(rgb\) or as grayscale images \(gray\).
We used our evaluation subset of the DIV2K dataset to test model responses on non-baked good images,
and reported the average false positives per image \(FP_N\) at a minimum confidence threshold of \(0.10\).
Table \ref{tab:ablation_data} shows model performances when training on different training data.
\begin{table}
	\setlength{\tabcolsep}{1.1pt}
	\renewcommand{\arraystretch}{1.1}
	\centering	
	\caption{Ablation study on training data.}
	\begin{tabular}{lccccc}
	\hline
	\textbf{Training set}		  & \(\bm{AP_{50}^{val}}\) & \(\bm{AP_{50}^{test}}\) & \(\bm{AP_{50}^{test_r}}\) & \(\bm{AP_{50}^{test_i}}\) & \(\bm{FP_N}\) \\ 
	\hline
	\hline
	\(\bm{train_{b}^{rgb}}\)	  &  \(13.4\%\)    & \(15.4\%\)   & \(13.7\%\)    & \(17.5\%\)     &  \(0.21\)                       \\
	\(\bm{train_{b,s}^{rgb}}\) 	  & \(46.6\%\)     & \(74.8\%\)   & \(79.5\%\)    & \(75.2\%\)     &  \(5.52\)                       \\
	\(\bm{train_{b,s,n}^{rgb}}\)  & \(59.8\%\)     & \(78.8\%\)   & \(83.5\%\)    & \(77.9\%\)     &  \(0.15\)                       \\
	\hline
	\hline
	\(\bm{train_{b}^{gray}}\)	  & \(19.1\%\)     & \(18.3\%\)   & \(20.1\%\)    & \(19.7\%\)     &  \(0.30\)                       \\
	\(\bm{train_{b,s}^{gray}}\)   & \(87.8\%\)     & \(80.5\%\)   & \(82.6\%\)    & \(82.7\%\)     &  \(3.91\)                       \\
	\(\bm{train_{b,s,n}^{gray}}\) & \(89.5\%\)     & \(84.7\%\)   & \(87.9\%\)    & \(83.8\%\)     &  \(0.02\)                       \\
	\hline 
	\end{tabular}
	\label{tab:ablation_data}
\end{table}
%
%\newline
%\newline
%
Our results indicate that our data synthesis was not only successful but essential in improving model performance. 
Furthermore, our results validate operating with grayscale images, as each model trained on grayscale images outperforms its counterpart trained on colored images.
This is likely due to the increased model robustness to different lighting conditions.
%
%It is noteworthy that models trained on colored images perform considerably better on the test set compared to the validation set, even though it was intentionally designed to be more challenging.
%
%This can be attributed to the challenging lighting conditions in our validation set.
%
Adding training images with random backgrounds significantly reduced \(FP_N\) and improved model performances. %, likely due to the increased diversity of synthetic images.
%
%Further analysis of the results shows that models perform the worst on \(test_u\), which closely simulated our use case by overcrowding the images with baked goods.
%
%This suggests that crowded images pose a greater challenge to the models compared to factors such as perspective or lighting conditions.
%
%The model trained on \(train_{b,s,n}^{gray}\) achieved a performance of \(74.8 AP_{50}\) on test set \(test_u\).
%
\subsubsection{Model Scales}
\label{sec:sec:ablation_scale}
Since our application does not have to operate in real-time, we do not emphasize image processing speed.
This allows us to run large models independently of the hardware on which they are deployed.
Table \ref{tab:ablation_scale} shows YOLOv8 model performances at varying scales.
\begin{table}
	\setlength{\tabcolsep}{1.1pt}
	\renewcommand{\arraystretch}{1.1}
	\centering
	\caption{Ablation study on model scale.}
	\begin{tabular}{lccccc}
		\hline
		\textbf{Model} & \textbf{\#Param} &  \(\bm{AP_{50}^{val}}\) & \(\bm{AP_{50}^{test}}\) & \(\bm{AP_{50}^{test_r}}\) & \(\bm{AP_{50}^{test_i}}\) \\ 
		\hline
		\hline
		\textbf{YOLOv8-N}	 & 3.2M		      & \(82.0\%\)                           & \(76.5\%\)                  & \(84.4\%\)                          & \(81.2\%\)            \\
		\textbf{YOLOv8-S}    & 11.2M		  & \(86.6\%\)                           & \(83.6\%\)                  & \(89.2\%\)                          & \(74.6\%\)            \\
		\textbf{YOLOv8-M}    & 25.9M		  & \(87.4\%\)                           & \(89.0\%\)                  & \(88.0\%\)                          & \(88.1\%\)            \\
		\textbf{YOLOv8-L}    & 43.7M		  & \(89.5\%\)                           & \(84.7\%\)                  & \(87.9\%\)                          & \(83.8\%\)            \\
		\textbf{YOLOv8-X}	 & 68.2M		  & \(89.3\%\)                           & \(86.4\%\)                  & \(89.4\%\)                          & \(85.5\%\)            \\
		\hline 
	\end{tabular}
	\label{tab:ablation_scale}
\end{table}
%
%\newline
%\newline
%
Our resuls indicate that YOLOv8-M achieves the best trade-off between performance and processing speed when models are trained on grayscale images with a maximum length of \(640px\).
Further increasing model scale results in, at best, marginal performance boosts.
\subsubsection{Online Image Augmentation}
\label{sec:sec:ablation_augmentation}
To verify the effectiveness of our online image augmentation pipeline, we trained our baseline model with:
no additional image augmentation \(None\),
the baseline augmentation pipeline \(BL_{0.01}\),
and our dropout image augmentation pipeline \(DO_{0.04}\).
Top half of Table \ref{tab:ablation_augmentation} shows the performances of our baseline model trained with or without our online image augmentation pipelines.
\begin{table}
	\setlength{\tabcolsep}{1.1pt}
	\renewcommand{\arraystretch}{1.1}
	\centering
	\caption{Ablation study on online image augmentation pipelines. Top half trained on our full training set \(train_{b,s,n}\). Bottom half trained on our reference training set \(train_{nA}\).}
	\begin{tabular}{llcccc}
		\hline
		\textbf{Aug} \phantom{0000} &\textbf{Training set}&  \(\bm{AP_{50}^{val}}\) & \(\bm{AP_{50}^{test}}\) & \(\bm{AP_{50}^{test_r}}\) & \(\bm{AP_{50}^{test_i}}\) \\ 
		\hline
		\hline
		\(\bm{None}\)	 		& \(train_{b,s,n}\)  & \(87.0\%\)                           & \(85.1\%\)                   & \(87.9\%\)                         &  \(85.1\%\)                     \\
		\(\bm{BL_{0.01}}\)  	& \(train_{b,s,n}\)  & \(89.5\%\)                           & \(85.2\%\)                   & \(87.4\%\)                         &  \(85.5\%\)                     \\
		\(\bm{DO_{0.04}}\)  	& \(train_{b,s,n}\)  & \(89.5\%\)                           & \(84.7\%\)                   & \(87.9\%\)                         &  \(83.8\%\)                     \\
		\hline
		\hline
		\(\bm{None}\)	 		&  \(train_{nA}\)    & \(79.6\%\)                            & \(72.8\%\)                    & \(67.8\%\)                          &  \(77.2\%\)                      \\
		\(\bm{BL_{0.01}}\)  	&  \(train_{nA}\)    & \(80.5\%\)                            & \(81.7\%\)                    & \(85.8\%\)                          &  \(81.3\%\)                      \\
		\(\bm{DO_{0.04}}\)  	&  \(train_{nA}\)    & \(85.3\%\)                            & \(84.3\%\)                    & \(88.1\%\)                          &  \(83.4\%\)                      \\
		\hline 
	\end{tabular}
	\label{tab:ablation_augmentation}
\end{table}
%
%\newline
%\newline
%
Our results indicate that applying our online image augmentation pipelines during model training leads to negligible performance improvements on the validation and test sets.
We tested whether this lack of improvement was caused by the augmentations applied in our image synthesis (Section \ref{sec:sec:image_composition}),
by creating a reference training set \(train_{nA}\) generated using our Copy-Paste augmentation pipeline, albeit limited to basic spatial-level augmentations: rotating and scaling of the individual baked goods.
\(train_{nA}\) comprises \(4000\) synthetic images with a synthetic background and \(2000\) images with a random background.
Bottom half of table \ref{tab:ablation_augmentation} shows the performances of our baseline model trained on reference training set \(train_{nA}\) with or without our online image augmentation pipelines.
%
%\newline
%\newline
%
Using our online image augmentation pipelines when training on \(train_{nA}\) improves performance considerably.
Although \(train_{nA}\) comprises a greater diversity of synthetic images, \(6000\) unique images were synthesized as opposed to \(4000\), training on \(train_{nA}\) results in a nearly universal performance drop.
This further validates the additional transformation applied in our image synthesis, and confirms that the rotating and scaling of synthetic images is essential to increase model performance.
\subsubsection{Image Sizes}
\label{sec:sec:image_size}
To standardize the image size and maintain a consistent scale across all images,
baked good images were resized such that the longest side is at most \(640px\).
We adopted this image size to facilitate a direct comparison with other object detection models and studies, and to enhance transfer learning, since YOLOv8 models are pre-trained on images of that particular scale.
To test whether this image size is a good fit for our detection task, we trained our baseline model on larger image sizes.
%
%Additionally, we trained the YOLOv8-X model on the larger image sizes to assess the potential benefits of larger image sizes for larger models.
%
Table \ref{tab:ablation_image_size} shows model performances when training with various image sizes.
\begin{table}
	\setlength{\tabcolsep}{1.1pt}
	\renewcommand{\arraystretch}{1.1}
	\centering	
	\caption{Ablation study on image size.}
	\begin{tabular}{llcccc}
		\hline
		\textbf{Model} \phantom{000000} & \textbf{Size} &  \(\bm{AP_{50}^{val}}\) & \(\bm{AP_{50}^{test}}\) & \(\bm{AP_{50}^{test_r}}\) & \(\bm{AP_{50}^{test_i}}\) \\ 
		\hline
		\hline
		\textbf{YOLOv8-L}     & 640		      & \(89.5\%\)        & \(84.7\%\)       & \(87.9\%\)       & \(83.8\%\)                     \\		
		\textbf{YOLOv8-L}     & 960	          & \(89.0\%\)        & \(88.5\%\)       & \(91.7\%\)       & \(87.5\%\)                     \\		
		\textbf{YOLOv8-L}     & 1280	      & \(90.3\%\)        & \(89.6\%\)       & \(92.5\%\)       & \(88.9\%\)                     \\
		\hline
		\hline
		\textbf{YOLOv8-X}     & 640		      & \(89.3\%\)        & \(86.4\%\)       & \(89.4\%\)       & \(85.5\%\)                     \\
		\textbf{YOLOv8-X}     & 960			  & \(87.3\%\)        & \(88.1\%\)       & \(91.8\%\)       & \(87.7\%\)                     \\
		\textbf{YOLOv8-X}     & 1280		  & \(91.7\%\)        & \(89.1\%\)       & \(92.7\%\)       & \(88.3\%\)                     \\
		\hline 
	\end{tabular}
	\label{tab:ablation_image_size}
\end{table}
%
%\newline
%\newline
%
Increasing the image size leads to considerable model performance improvements.
YOLOv8-L and YOLOv8-X both achieve the best performance when operating on images with a maximum size of \(1280px\).
%
%YOLOv8-X achieved the overall better performance, which indicates that larger models benefit more from larger images.
%
\subsection{Model Comparison}
\label{sec:sec:model}
To support this study's focus on the YOLOv8 model, we trained state-of-the-art object detection models YOLOv5 and defDETR on our detection task and compared performances.
Model scales were chosen to ensure that the models have roughly the same number of trainable parameters.
Table \ref{tab:ablation_image_size} shows the performances of the evaluated object detection models.
\begin{table}
	\setlength{\tabcolsep}{1.1pt}
	\renewcommand{\arraystretch}{1.1}
	\centering	
	\caption{Object detection model comparison.}
	\begin{tabular}{lccccc}
		\hline
		\textbf{Model} & \textbf{\#Param} & \(\bm{AP_{50}^{val}}\) & \(\bm{AP_{50}^{test}}\) & \(\bm{AP_{50}^{test_r}}\) & \(\bm{AP_{50}^{test_i}}\) \\  
		\hline
		\hline
	    %\textbf{YOLOv7}	  & 36.9M		      & \(tbd\)                            & \(tbd\)                  & \(tbd\)                          &  \(tbd\)                       \\		
		\textbf{defDETR}      & 40.0M		      & \(81.1\%\)                         & \(80.9\%\)               & \(88.8\%\)                       &  \(78.9\%\)                    \\
		\textbf{YOLOv5-L}	  & 46.5M		      & \(86.2\%\)                         & \(84.8\%\)               & \(87.1\%\)                       &  \(84.4\%\)                    \\
		\textbf{YOLOv8-L}     & 43.7M		      & \(89.5\%\)                         & \(84.7\%\)               & \(87.9\%\)                       &  \(83.8\%\)                    \\
		\hline 
	\end{tabular}
	\label{tab:ablation_models}
\end{table}
%
%\newline
%\newline
%
Each tested object detection model achieved an \(AP_{0.5}\) of over \(80\%\) on both our validation and test sets.
YOLOv8 achieved the best overall performance on our detection task, validating the claim of improved performance made by its developers \cite{YOLOv8}.
Its predecessor YOLOv5 achieved the second-best overall performance, while the defDETR model achieved the worst overall performance on our detection task.
Potential reasons for the lower performance of defDETR, are:
(1) defDETR employs less elaborate data augmentation by default compared to the tested YOLO models.
(2) defDETR drastically increased convergence speed compared to its predecessor DETR, it may still require more training data than the CNN-based YOLO models.
\subsection{Analysis of the Best Performing Model}
\label{sec:sec:top_model_results}
\begin{figure}
	\begin{center}
		\includegraphics[trim={0cm 0cm 2cm 0cm},clip,width=0.98\linewidth]{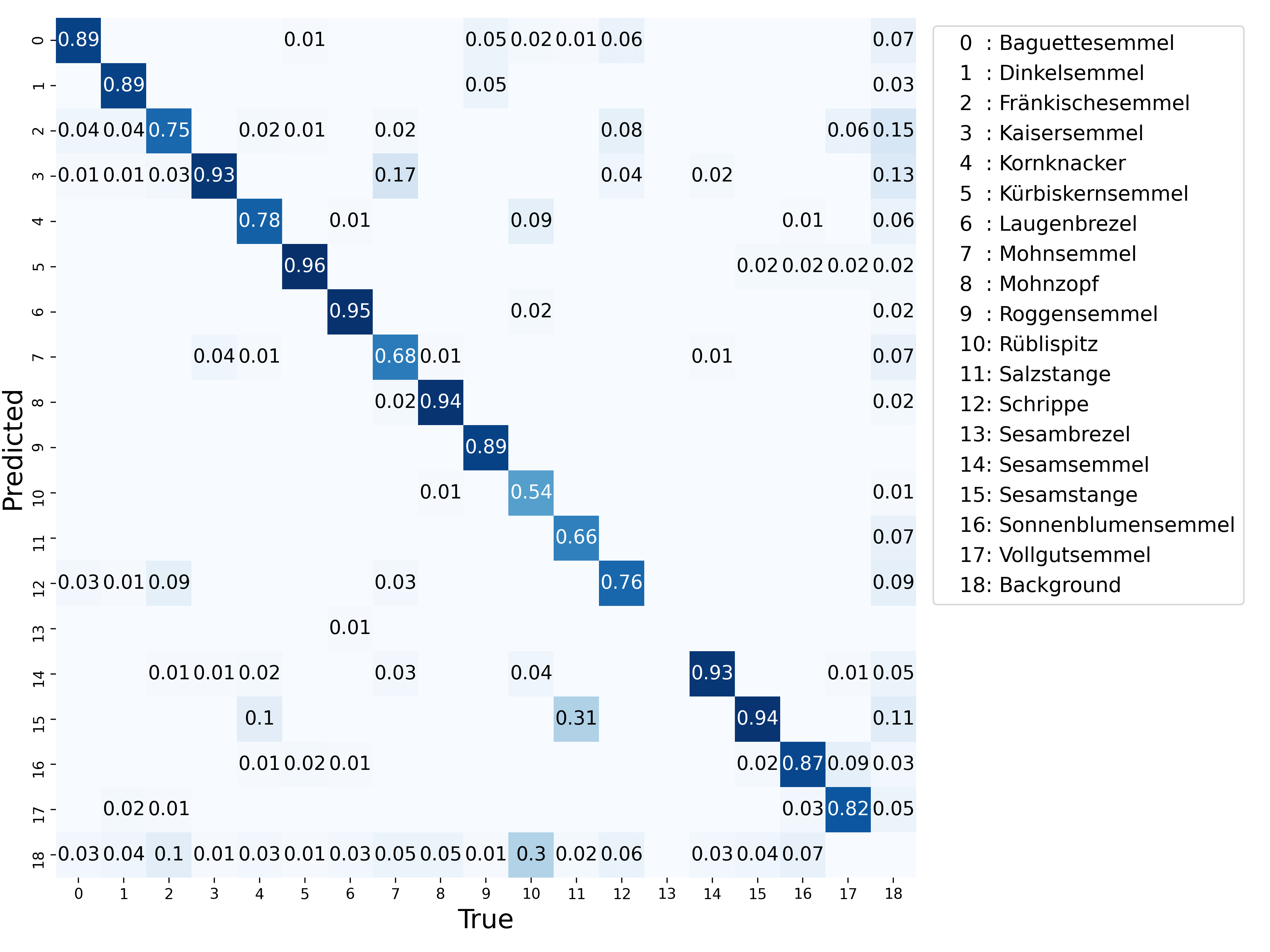}
	\end{center}
	\caption{Confusion Matrix of our best model's test set predictions, at minimum confidence and  \(IoU\) thresholds of \(0.25\) and \(0.45\), respectively.}
	\label{fig:CM}
\end{figure}

Our overall best performing model, a YOLOv8-X model, achieved a \(AP_{0.5}\) of \(89.1\%\) on our test set.
The model was trained on our full training set \(train_{b,s,n}^{gray}\), converted into grayscale images and resized such that the longest side is at most \(1280px\).
During training, we used our complementary image augmentation pipeline, \(DO_{0.05}\).
%
%Figure \ref{fig:PR} shows the mean and class-specific Precision-Recall curves of our best model's test set predictions.
%
Figure \ref{fig:CM} shows the confusion matrix of our best model's test set predictions, at minimum confidence and \(IoU\) thresholds of \(0.25\) and \(0.45\), respectively.
%
%\newline
%\newline
%
Despite our efforts to reduce \(FP_N\), the confusion matrix shows that misclassifying the background as baked goods and vice versa remains one of the most prevalent prediction errors.
%
%Furthermore, it shows that our model's ability to detect and classify baked goods varies depending on their type.
%
Rüblispitz (spelt breads with beet syrup), Salzstange (salt breadsticks), and Mohnsemmel (poppy seed bread buns) are especially challenging for our model.
%
%\begin{figure}
%	\begin{center}
%		\includegraphics[trim={0cm 0cm 2cm 0cm},clip,width=0.98\linewidth]{PR_curve.png}
%	\end{center}
%	\caption{Mean and class-specific Precision-Recall curves of our best model's test set predictions.}
%	\label{fig:PR}
%\end{figure}
%
%Rüblispitz is frequently missed by our model, as it often mistakes it for the background. 
%
%This behavior is reflected in the Precision-Recall curve, where our model achieved the lowest class-specific \(AP_{0.5}\) of \(69.4\%\) for Rüblispitzen.
%
%Our model achieves a class-specific \(AP_{0.5}\) of close to or well exceeding \(80\%\) for the remaining baked goods.
%
Salzstangen and Mohnsemmeln are detected badly by our model as they are often misclassified as other types of baked goods, namely Sesamstange (sesame breadstick) and Kaisersemmel (emperor bread bun).
The model frequently confuses these two pairs due to their similar shapes and colors, particularly when they are turned upside down.
%
%The remaining \(15\) less difficult baked goods are detected with an accuracy of at least \(75\%\) at the chosen thresholds.
%
%Rüblispitz's class-specific PR curve, exemplifies how the average precision \(AP\) is sensitive to the number of unique class-likelihood predictions, (Section \ref{sec:sec:average_precision}).
%
%Specifically, a single prediction resulting in a Recall and Precision between \((0.5,1.0)\) and  \((0.0,0.7)\), respectively, could significantly impact the class-specific \(AP_{0.5}\) for Rüblispitzen.
%

%% file: text/end.tex
\section{Conclusions}
\label{sec:conclusions}
In this study, we presented the Semmeldetector, an application of object detection model to detect, classify and count baked goods in images.
To train our models, we compiled a dataset comprising \(1151\) images that differentiates between \(18\) distinct baked goods.
%
%We limited our training set to images of individual baked goods to automatically annotate our images and rapidly scale our models.
%
We demonstrated how a Copy-Paste augmentation pipeline can be used to enrich small datasets, such as the one we used in this study, to facilitate model training.
%
%Our results demonstrated that the YOLOv8 is well-suited for our task.
%
Specifically, our overall best performing model, achieved an \(AP_{0.5}\) of \(89.1\%\) on our test set while operating on grayscale images.
Furthermore, our study highlighted the importance of incorporating random images in the training of object detection models deployed in real-world scenarios to mitigate  \(FP\).
We conducted an ablation study to test the impact of training data, model scale, online image augmentation pipeline, and image size on model performance.
Overall, our study highlights the potential of machine learning models to optimize the production of baked goods.
\section{Future Work}
\label{sec:future_work}
Our study could benefit from further research in the following areas:
(1) Expanding the scope of our model comparison to include YOLOv7 \cite{YOLOv7}, DETR \cite{DETR}, and Swin-T \cite{Swin-T} could provide further insight into the effectiveness of object detection models for our detection task.
(2) Exploring generative models for synthesizing images of baked goods could greatly benefit our models.
(3) Expanding our training data to a wider range of baked goods would bolster both performance and applicability of our models.
\section{Acknowledgments}
%
%We would like to thank someone (but this is a blind review).
We would like to thank Backhaus Müller, local Franconian bakery for their cooperation and insight.